\documentclass[10pt,twocolumn,letterpaper]{article}

\usepackage{booktabs}
\usepackage{makecell}
\usepackage{multirow} 
\usepackage[table,dvipsnames]{xcolor}
\definecolor{navyblue}{HTML}{0071BC}
\usepackage{tabularx}  
\usepackage{fontawesome5} 
\usepackage{multirow} 

\usepackage[pagenumbers]{cvpr} 

\definecolor{cvprblue}{rgb}{0.21,0.49,0.74}
\usepackage[pagebackref,breaklinks,colorlinks,allcolors=cvprblue]{hyperref}

\def\confName{CVPR}
\def\confYear{2027}

\title{EVAThink: Evidence Sufficiency-Guided Training for Adaptive Thinking in Video Understanding}

\author{
    Changti Wu\textsuperscript{\rm 1,\rm 2},
    Ruizhe Fan\textsuperscript{\rm 3},
    Bin Wang\textsuperscript{\rm 3},
    Lei Zhang\textsuperscript{\rm 1}\thanks{Corresponding author},
    Yuxiang Zhang\textsuperscript{\rm 3}\footnotemark[1]\\
    \textsuperscript{1}East China Normal University
    \textsuperscript{2}Zhongguancun Academy
    \textsuperscript{3}Huawei Technologies Co., Ltd
}

\begin{document}
\maketitle
\begin{abstract}
Under a fixed frame budget, video understanding with vision-language models (VLMs) depends on whether the sampled frames provide sufficient evidence for the question. Sufficient frame evidence supports direct answering, whereas less informative observations may benefit from reasoning that connects the available cues.
We introduce EVAThink, which enhances video understanding through evidence sufficiency-guided training for adaptive thinking.
First, Recoverable Evidence Gap Mining constructs evidence-dense and evidence-sparse video views that share the same question and answer under an identical frame budget.
It retains pairs where sparse views yield lower direct-answer accuracy than dense views but benefit from explicit reasoning, providing supervision for response-mode selection.
Second, Evidence-Guided Two-Stage Reinforcement Learning uses stage-specific rewards to guide training with evidence-based mode preferences and then refine response efficiency, while first-token advantage regularization mitigates mode collapse.
Trained on only 6.7K curated samples,
EVAThink consistently improves upon the base model across video understanding benchmarks under both 32- and 64-frame budgets, while outperforming strong baselines on average.
The code, model, and datasets are available at \href{https://github.com/ChangtiWu/EVAThink}{EVAThink}.
\end{abstract}

\section{Introduction}

Vision-language models (VLMs) have made substantial progress in video understanding, including video captioning, temporal grounding, and video question answering \cite{tang2025video,wu2025survey,zohar2025apollo,lian2026physbrain,wu2025dynasolidgeo,wu2026scalselect}.
In practice, these models often operate under a limited visual budget, accessing a sampled sequence of frames rather than the complete video.
Their answers therefore depend on the question-relevant information preserved in those frames.
We refer to this information as \emph{frame evidence}, which includes visual cues within individual frames and temporal relationships across frames.
With a fixed budget of uniformly sampled frames, some sampled sequences provide sufficient evidence for a reliable direct answer, while others offer only sparse coverage of question-relevant content, particularly when that content occupies a small fraction of the video.
We term the extent to which frame evidence supports reliable direct answering by the model \emph{evidence sufficiency}, an important consideration when allocating reasoning computation.

\begin{figure}[!t]
  \centering
  \includegraphics[width=\linewidth]{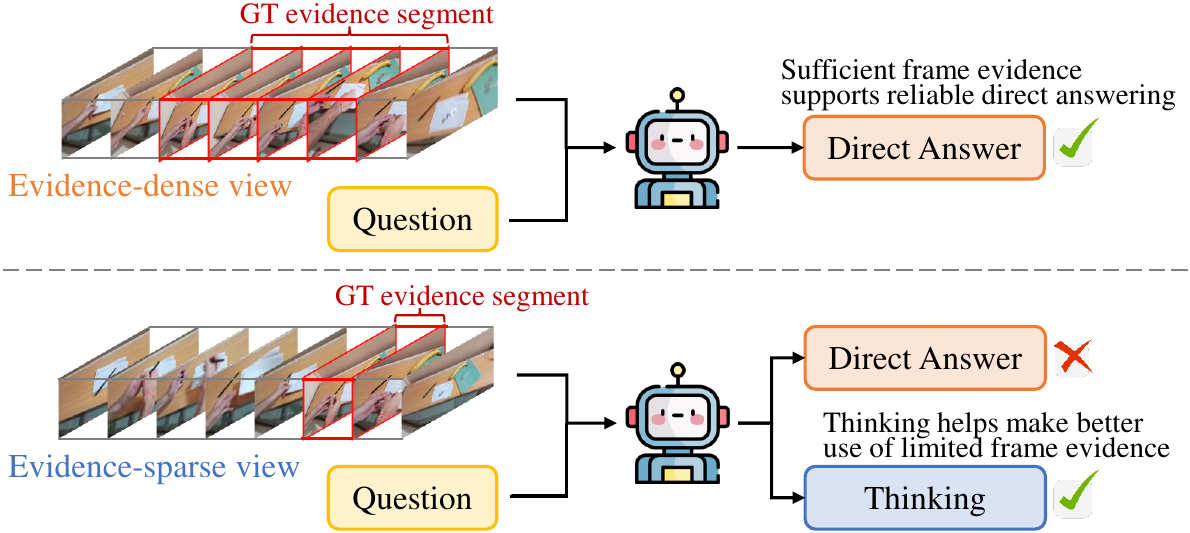}
  \caption{Motivation for evidence sufficiency-guided training for adaptive thinking.
For the same question and frame budget, sufficient frame evidence supports direct answering, while thinking can help make better use of limited frame evidence.}
   \label{fig:data_pool}
\end{figure}

Explicit reasoning can help models interpret frame evidence by connecting observations across frames.
However, when the evidence already supports a reliable direct answer, additional reasoning may yield little benefit while increasing generation cost and introducing unsupported inferences.
When useful cues are present but require further analysis, reasoning may instead improve the answer.
This motivates using evidence sufficiency to guide adaptive-thinking training, encouraging direct answering when frame evidence is sufficient and reasoning when a direct answer is less reliable but further analysis may help.

Recent work on adaptive reasoning for video understanding selects response modes using initial-answer confidence \cite{liu2026videoauto} or objectives that balance answer quality, including correctness and grounding, against generation cost \cite{tian2026adathinkv,kulkarni2026video}.
These approaches provide useful signals for mode selection given a visual input, but supervision across different evidence conditions remains less explored. Such supervision could use observed differences in direct-answer accuracy and reasoning gains to guide adaptive-thinking training.

Learning such a response-mode policy also poses challenges in training stability.
Prior studies report that adaptive reasoning training can prematurely favor one mode, limiting exploration of the other and potentially leading to mode collapse, in which the model consistently answers directly or always reasons \cite{fang2026thinkless,tu2026learning,liu2026videoauto}.
Existing methods can also require substantial training data, e.g., 83K samples for VideoAuto-R1~\cite{liu2026videoauto} and 140K for Video-FLAIR~\cite{kulkarni2026video}.
Some further rely on a supervised fine-tuning (SFT) cold start, adding the burden of preparing and annotating response demonstrations \cite{kulkarni2026video,tian2026adathinkv}.

To address these challenges, we introduce \textbf{EVAThink}, which learns adaptive thinking from differences in frame evidence sufficiency.
During data curation, our \textit{Recoverable Evidence Gap Mining} constructs evidence-contrastive pairs of dense and sparse video views that share the same question and answer under a fixed frame budget.
We retain pairs with higher direct-answer accuracy on the dense view, lower direct-answer accuracy on the sparse view, and a sufficient accuracy gain from explicit reasoning on the sparse view.
These pairs capture a recoverable evidence gap, linking differences in frame evidence sufficiency to the benefit of reasoning.
Building on these pairs, we develop \textit{Evidence-Guided Two-Stage Reinforcement Learning}, which requires no SFT cold start.
Using stage-specific rewards, the first stage guides response-mode training with evidence-based mode preferences from high-confidence pairs while rewarding answer correctness; the second removes view-dependent mode preferences and adds a small bonus for correct direct answers to improve efficiency.
First-token advantage regularization mitigates mode collapse.

With only 6.7K curated samples, EVAThink consistently improves over the base model across six video understanding benchmarks and outperforms a range of strong baselines overall.
Its adaptive thinking policy outperforms direct-only and think-only mode by 2.3 and 3.6 percentage points, respectively, while reducing average output length by approximately 62\% compared with think-only inference.
Our main contributions are as follows:
\begin{itemize}
    \item We propose \textit{Recoverable Evidence Gap Mining} to construct evidence-contrastive pairs comprising evidence-dense and evidence-sparse views of the same question-answer pair. These pairs link a decline in direct-answer accuracy on the sparse view to a gain from explicit reasoning, providing targeted supervision for evidence-guided adaptive thinking.

    \item We develop \textit{Evidence-Guided Two-Stage Reinforcement Learning} with stage-specific rewards that first guide training with evidence-based mode preferences and then refine response efficiency. The framework incorporates first-token advantage regularization to mitigate mode collapse.

    \item Using only 6.7K samples for training, EVAThink consistently improves over the base model across video understanding benchmarks and outperforms strong baselines overall under controlled visual budgets. Analysis shows that adaptive thinking achieves higher accuracy than both direct-only and think-only modes while substantially reducing output length relative to think-only mode.
\end{itemize}
\section{Related Work}
\subsection{Video Understanding}
Vision-language models (VLMs) have advanced video understanding through instruction tuning and temporal modeling~\cite{tang2025video,wu2025survey,zohar2025apollo,zhang2024llava,bai2025qwen25vltechnicalreport,bai2025qwen3,zhang2026progressive}.
Building on these capabilities, recent studies explore explicit reasoning and reinforcement learning (RL) for video tasks.
Video-R1~\cite{feng2026video} introduces temporal-aware rewards, while VideoRFT~\cite{wang2026videorft} combines reasoning supervision with RL.
Temporal-RLT~\cite{li2025reinforcement} and Video-RTS~\cite{wang2025video} investigate data-efficient RL, while LongVILA-R1~\cite{chen2026scaling} extends RL to long videos.
VITAL~\cite{zhang2026thinking}, LOVE-R1~\cite{fu2025love}, and VideoChat-R1.5~\cite{yan2026videochat} incorporate tool use or iterative perception into reasoning.
Complementary approaches enhance video understanding through dedicated frame-selection modules~\cite{hu2025m,han2025videoespresso}, explicit evidence-grounding mechanisms~\cite{huang2026video}, or agentic pipelines to guide further observation and retrieval~\cite{wang2026active,yan2026reveal}.
These designs incur additional inference-time overhead for selection, grounding, or retrieval, while leaving native adaptive thinking less explored.
EVAThink uses RL with paired evidence conditions to train the VLM itself to choose when to reason, without external tools or additional retrieval at inference.

\subsection{Adaptive Thinking}
Adaptive thinking adjusts reasoning computation to balance answer quality and efficiency.
For language models, existing approaches include response-mode selection~\cite{fang2026thinkless,zhang2025adaptthink}, multi-stage reinforcement learning~\cite{tu2026learning}, accuracy-aware length rewards~\cite{zhang2025continue}, and confidence-guided reflection~\cite{wan2025adapthink}.
For video understanding, VideoAuto-R1~\cite{liu2026videoauto} uses initial-answer confidence to determine whether further reasoning is needed.
AdaThinkV~\cite{tian2026adathinkv} estimates the utility of explicit reasoning by balancing accuracy gains against generation cost, while Video-FLAIR~\cite{kulkarni2026video} compares reasoning modes based on correctness, grounding, and cost.
While these methods adapt reasoning to individual inputs, explicit supervision across paired evidence conditions for the same question remains less explored.
EVAThink constructs paired video views that share the same question and answer but differ in the availability of question-relevant evidence under a fixed frame budget.
It uses differences in direct-answer accuracy and reasoning gains across these views to guide adaptive-thinking training.

\begin{figure*}[!t]
  \centering
  \includegraphics[width=\linewidth]{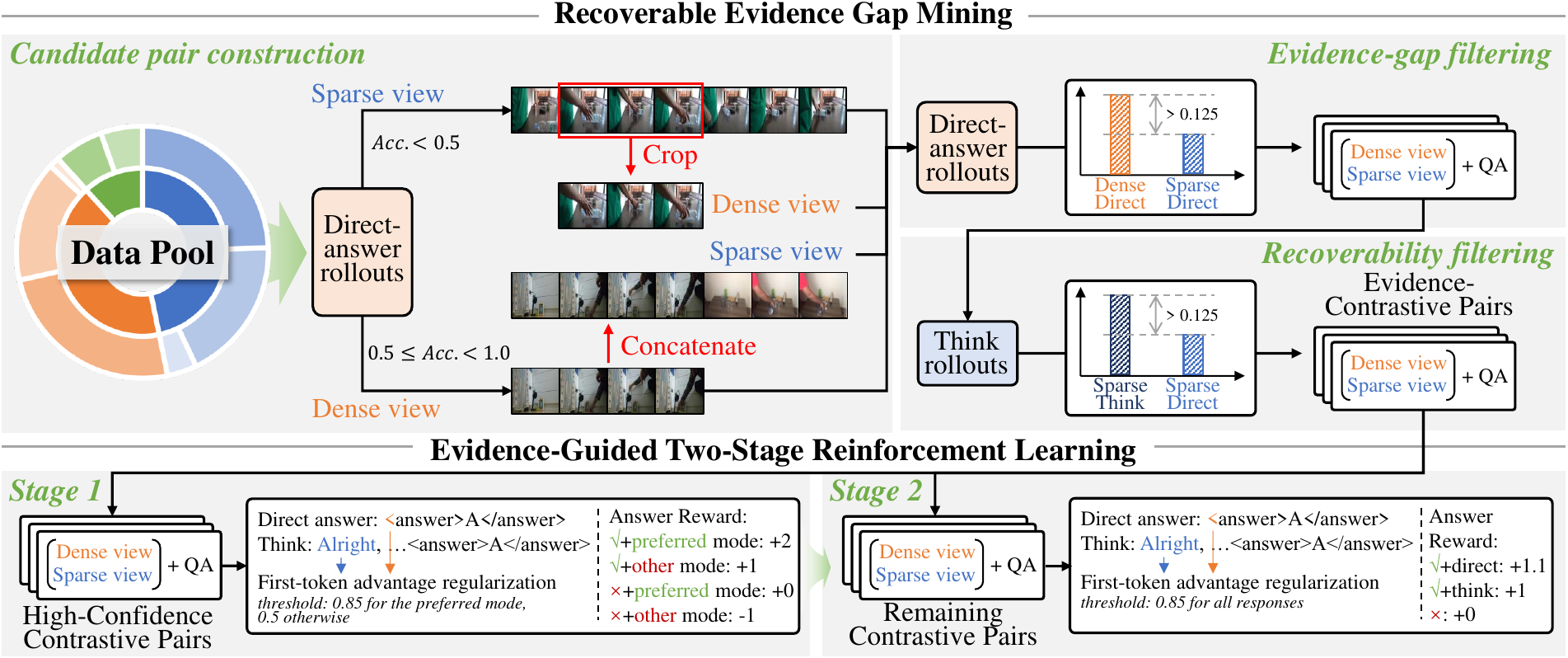}
  \caption{Overview of EVAThink. Recoverable Evidence Gap Mining constructs the evidence-contrastive pairs through video cropping or concatenation, retaining pairs with a direct-answer accuracy gap and a reasoning gain on the sparse view. Evidence-Guided Two-Stage Reinforcement Learning first trains response-mode selection using preferences derived from evidence-contrastive pairs, then refines response efficiency on the remaining pairs, using stage-specific rewards and first-token advantage regularization.}
   \label{fig:overview}
\end{figure*}

\section{EVAThink}
We propose EVAThink, which enhances video understanding through evidence sufficiency-guided training for adaptive thinking. As illustrated in Fig.~\ref{fig:overview}, EVAThink comprises data curation and policy training, where Recoverable Evidence Gap Mining constructs evidence-contrastive pairs for Evidence-Guided Two-Stage Reinforcement Learning. Stage 1 trains response-mode selection using preferences derived from evidence-contrastive pairs, while Stage 2 refines response efficiency. Both stages use stage-specific rewards and first-token advantage regularization.

\begin{figure}[!th]
  \centering
  \includegraphics[width=0.9\linewidth]{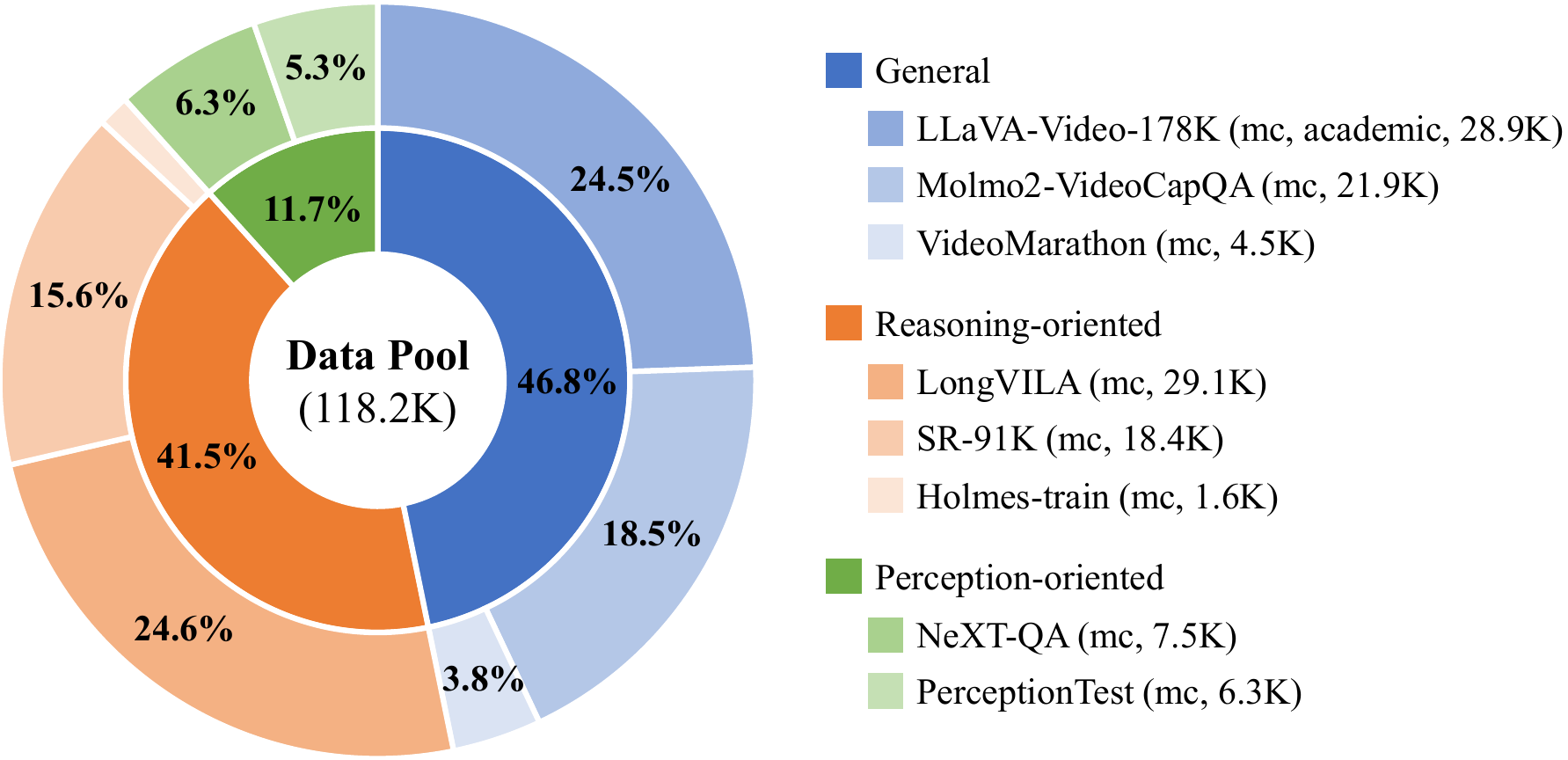}
  \caption{Data pool.}
   \label{fig:data_pool}
\end{figure}

\paragraph{Data Collection.}
We first construct a candidate pool of 118.2K multiple-choice video-question samples from eight existing datasets, covering diverse video understanding tasks. As shown in Fig.~\ref{fig:data_pool}, we organize the pool into three groups based on their primary focus: \textit{general video understanding} (55.3K, 46.8\%), including LLaVA-Video-178K \cite{zhang2024llava}, Molmo2-VideoCapQA \cite{clark2026molmo2}, and VideoMarathon \cite{lin2026unleashing}; \textit{reasoning-oriented video understanding} (49.1K, 41.5\%), including LongVILA \cite{chen2025longvila}, SR-91K \cite{ouyang2025spacer}, and Holmes-train \cite{cheng2026video}; and \textit{perception-oriented video understanding} (13.8K, 11.7\%), including NeXT-QA \cite{xiao2021next} and PerceptionTest \cite{patraucean2023perception}. This pool provides the candidate samples for subsequent Recoverable Evidence Gap Mining.
See Appendix~\ref{app:data_pool} for more details.

\subsection{Recoverable Evidence Gap Mining}
\label{sec:data_curation}

Here, we construct evidence-contrastive pairs comprising an \emph{evidence-dense view} and an \emph{evidence-sparse view} that share the same question and answer. The two views differ in the fraction of video duration occupied by the evidence segment relevant to the question: this fraction is larger in the dense view and smaller in the sparse view. Under a fixed budget of uniformly sampled frames, the dense view is therefore expected to provide greater coverage of question-relevant evidence.
We retain pairs for which direct-answer accuracy decreases from the dense to the sparse view, while explicit reasoning improves accuracy on the sparse view.
These pairs provide supervision for adaptive-thinking training across different evidence conditions.

Starting from the candidate data pool, we construct such pairs in three steps (see Appendix \ref{app:impl_detail_0} for more details).

\noindent\textbf{1) Candidate pair construction.}
Throughout data curation, the target model receives 32 frames uniformly sampled from each video.
For each evaluated view and response mode, we estimate answer accuracy using eight rollouts.
We first estimate direct-answer accuracy, $A_{\mathrm{dir}}$, using eight rollouts per sample. We discard samples answered correctly in all eight rollouts, as they are likely too easy to provide useful signals for subsequent reinforcement learning (RL).
For samples with $A_{\mathrm{dir}} < 0.5$, we use the original video as the sparse view and construct the dense view by cropping around an evidence segment annotated by a stronger model given the video, question, and answer.
We extend the crop to include 2--3 sampled frames before and after the annotated segment, using their timestamps in the original video's uniformly sampled 32-frame sequence.
For the remaining samples, with $0.5 \leq A_{\mathrm{dir}} < 1$, we use the original video as the dense view and construct the sparse view by concatenating another video from the same data source, placing it randomly before or after the original video.
For questions involving temporal intervals, we restrict cropping and concatenation to the portion after the annotated segment to preserve the original timestamps.

\noindent\textbf{2) Evidence-gap filtering.}
For each candidate pair, we evaluate the newly constructed view using eight direct-answer rollouts and retain pairs satisfying
\begin{equation}
\Delta_{\mathrm{evi}}
=
A_{\mathrm{dir}}^{\mathrm{dense}}
-
A_{\mathrm{dir}}^{\mathrm{sparse}}
> 0.125,
\end{equation}
where $A_{\mathrm{dir}}^{\mathrm{dense}}$ and $A_{\mathrm{dir}}^{\mathrm{sparse}}$ denote the direct-answer accuracies on the dense and sparse views, respectively.
We use this accuracy gap as an empirical proxy for the difference in how reliably the two views support direct answering.

\noindent\textbf{3) Recoverability filtering.}
For each pair retained after evidence-gap filtering, we prompt the model to use explicit reasoning on the sparse view and retain pairs satisfying
\begin{equation}
\Delta_{\mathrm{think}}
=
A_{\mathrm{think}}^{\mathrm{sparse}}
-
A_{\mathrm{dir}}^{\mathrm{sparse}}
> 0.125,
\end{equation}
where $A_{\mathrm{think}}^{\mathrm{sparse}}$ denotes the reasoning accuracy on the sparse view.
These pairs exhibit a recoverable evidence gap: direct answering is less accurate on the sparse view than on the dense view, but explicit reasoning improves accuracy on the sparse view using the same sampled frames.
Here, recoverability refers to improved answer accuracy, not the recovery of unobserved visual information.
The resulting evidence-contrastive pairs are used for subsequent training.

\subsection{Evidence-Guided Two-Stage Reinforcement Learning}

To explicitly anchor the two response modes, we specify distinct response formats for two response modes during training.
Specifically, we append a training prompt to each question that instructs the model to answer directly when the visual evidence clearly supports an answer, and to use explicit reasoning when further analysis is needed.
The prompt requires each response to follow one of two formats, with distinct openings identifying the selected mode:
\begin{center}
\small
\begin{tabular}{@{}l@{}l@{}}
\text{Direct answer: } &
\texttt{<answer>[Answer]</answer>} \\
\text{Think: } &
\texttt{Alright,[Reasoning]} \\
&\texttt{<answer>[Answer]</answer>}
\end{tabular}
\end{center}
This prompt elicits both response modes during initial rollouts, enabling RL training without an SFT cold start, avoiding the need to curate supervised response demonstrations.
The distinct openings also support first-token advantage regularization, as described below.
The full training prompt is provided in Table~\ref{tab:prompt_training} in the Appendix \ref{app_training_prompt}.

Building on the Group Relative Policy Optimization (GRPO) \cite{shao2024deepseekmath,guo2025deepseek}, we develop a two-stage RL framework that first guides training with evidence-based mode preferences and then refines response efficiency.
Our training incorporates two key designs:
(i) stage-specific rewards that guide mode selection in Stage~1 and refine response-efficiency in Stage~2;
(ii) a first-token advantage regularization mechanism to mitigate entropy collapse during training and discourage premature concentration on a single response mode;
To maintain balanced exposure to paired evidence conditions, we additionally use paired sampling, including the dense and sparse views of each selected pair in the same training batch.
Besides, following DAPO \cite{yu2026dapo}, we omit the KL-divergence penalty and adopt token-level mean loss aggregation throughout both training stages.

In both training stages, we uniformly sample 32 frames from each video, consistent with data curation.
We next describe the reward design and first-token advantage regularization for each stage.

\noindent\textbf{Stage 1 RL.}
In this stage, we aim to use high-confidence evidence-contrastive pairs to guide response-mode training.
From the evidence-contrastive pairs obtained through Recoverable Evidence Gap Mining, we select high-confidence pairs satisfying $A_{\mathrm{dir}}^{\mathrm{dense}} \geq 0.75$ and $\Delta_{\mathrm{think}} \geq 0.5$.
These pairs motivate training preferences for direct answering on dense views and explicit reasoning on sparse views, based on the evaluated rollout accuracies.

\emph{Reward design.}
Let $x=(v,q)$ denote a video-question input, $z\in\{\mathrm{dense},\mathrm{sparse}\}$ its view type, and $o_i$ a sampled response with mode $m_i\in\{\mathrm{direct},\mathrm{think}\}$.
The preferred response mode is defined as
\begin{equation}
m^\star(z)=
\begin{cases}
\mathrm{direct}, & z=\mathrm{dense},\\
\mathrm{think}, & z=\mathrm{sparse}.
\end{cases}
\end{equation}

The answer reward jointly encourages answer correctness and alignment with the preferred mode:
\begin{equation}
R_i^{\mathrm{ans}} =
\begin{cases}
+2, & \text{correct answer},\ m_i=m^\star(z),\\
+1, & \text{correct answer},\ m_i\neq m^\star(z),\\
0,  & \text{incorrect answer},\ m_i=m^\star(z),\\
-1, & \text{incorrect answer},\ m_i\neq m^\star(z).
\end{cases}
\end{equation}
The format reward penalizes violations of the response formats specified above:
\begin{equation}
R_i^{\mathrm{fmt}} =
\begin{cases}
0,  & \text{valid response format},\\
-1, & \text{otherwise}.
\end{cases}
\end{equation}
The total Stage 1 reward is
\begin{equation}
R_i^{(1)} = R_i^{\mathrm{ans}} + R_i^{\mathrm{fmt}}.
\label{eq:stage1_reward}
\end{equation}

\emph{First-token advantage regularization.}
For responses following the prescribed formats, the distinct first tokens (\texttt{<} and \texttt{Alright} for Qwen3-VL) identify the response mode.
We regularize the first-token advantage to mitigate entropy collapse and premature concentration on a single mode.
Let $A_i$ denote the GRPO group-relative advantage of response $o_i$, and let $p_i=\pi_{\mathrm{old}}(o_{i,1}\mid x)$ be the probability of its first generated token.
The regularized token-level advantage is
\begin{equation}
\widetilde A_{i,t}
=
A_i
-
\mathbb{I}[t=1]\,
\beta\max(\log p_i-\log c_i,0),
\label{eq:first_token_regularization}
\end{equation}
where $c_i$ is a probability threshold and $\beta$ controls the regularization strength.
This adjustment reduces the first-token advantage when $p_i$ exceeds $c_i$, leaving all subsequent token advantages unchanged.
In Stage~1, we set $c_i=0.85$ when $m_i=m^\star(z)$ and $0.50$ otherwise, applying stronger regularization to confident selections of the non-preferred mode.
We use $\beta=5$ in both stages.

For each input $x$, we sample $G$ responses $\{o_i\}_{i=1}^{G}$ from $\pi_{\mathrm{old}}$ and optimize the clipped surrogate objective with token-level mean aggregation and no KL penalty:
\begin{equation}
\begin{aligned}
\mathcal{J}(\theta)
&=
\mathbb{E}_{
x\sim\mathcal{D},
\{o_i\}_{i=1}^{G}\sim\pi_{\mathrm{old}}(\cdot\mid x)
}
\Bigg[
\frac{1}{\sum_{i=1}^{G}|o_i|}
\sum_{i=1}^{G}\sum_{t=1}^{|o_i|}
\\
&
\min\Big(
\rho_{i,t}(\theta)\widetilde A_{i,t},
\operatorname{clip}\big(
\rho_{i,t}(\theta),1-\varepsilon,1+\varepsilon
\big)\widetilde A_{i,t}
\Big)
\Bigg],
\end{aligned}
\label{eq:policy_objective}
\end{equation}
where $\mathcal{D}$ is the training set for the current stage, $\varepsilon$ is the clipping parameter, and
$
\rho_{i,t}(\theta)
=
\frac{
\pi_\theta(o_{i,t}\mid x,o_{i,<t})
}{
\pi_{\mathrm{old}}(o_{i,t}\mid x,o_{i,<t})
}.
$

\noindent\textbf{Stage 2 RL.}
This stage aims to refine response efficiency while maintaining answer accuracy.
Starting from the Stage~1 policy, we continue training on the remaining evidence-contrastive pairs after excluding the high-confidence pairs used in Stage~1.

\emph{Reward design.}
As Stage~1 updates the policy, the view-based mode preferences inferred during data curation may become less reliable for the remaining, lower-confidence pairs.
We therefore remove the view-dependent mode preference used in Stage~1 and introduce a small bonus for correct direct answers:
\begin{equation}
R_i^{\mathrm{ans}} =
\begin{cases}
+1.1, & \text{correct answer},\ m_i=\mathrm{direct},\\
+1,   & \text{correct answer},\ m_i=\mathrm{think},\\
0,    & \text{incorrect answer}.
\end{cases}
\end{equation}
Building on evidence sufficiency-guided data curation and Stage 1 training, this reward further refines response-mode selection through answer correctness, with a small bonus favoring correct direct answers.
Using the same format reward as in Stage~1, the total Stage~2 reward is
\begin{equation}
R_i^{(2)} = R_i^{\mathrm{ans}} + R_i^{\mathrm{fmt}}.
\label{eq:stage2_reward}
\end{equation}

\emph{First-token advantage regularization.}
We retain the regularization in Eq.~\eqref{eq:first_token_regularization}, using a shared threshold $c_i=0.85$ for both response modes and keeping $\beta=5$.
This removes the mode-dependent thresholds used in Stage~1 while continuing to discourage premature concentration on either mode.
We optimize the policy using the same objective in Eq.~\eqref{eq:policy_objective}.

\begin{table*}[!thp]
\centering
\renewcommand{\arraystretch}{1.1}
\setlength\tabcolsep{1em} 
\resizebox{\linewidth}{!}
{
\begin{tabular}{l|c|cccccc|c}
\toprule
\textbf{Model} & \textbf{Train. Size} & \textbf{Video-MME} & \textbf{MVBench} &	\textbf{LongVideoBench} & \textbf{MMVU(mc)} & \textbf{Video-MMMU} & \textbf{VSI-Bench} & \textbf{Avg.} \\
\midrule
\multicolumn{7}{l}{\textit{\textbf{Results reported under the respective evaluation protocols (different visual budgets and evaluation settings)}}} \\
Temporal-RLT & 32K & 57.6 & 68.1 & - & 65.0 & - & - & - \\
Video-RTS & 6K & 63.0 & - & 56.6 & 66.4 & 52.7 & - & -\\
VITAL & 182K & 64.1 & - & - & 68.7 & 54.2 & 41.8 & - \\
LongVILA-R1 & 206K & 65.1 & 67.6 & 58.0 & - & - & - & -\\
LOVE-R1 & 192.5K & 66.2 & 66.6 & 60.1 & - & - & -& - \\
VideoChat-R1.5 & 80K & 65.2 & 70.6 & 61.4 & - & 49.6 & - & -\\
AdaThinkV$^{\dagger}$ & 40.7K & 72.1 & 72.5 & 68.1 & 71.4 & 65.3 & -& - \\
Video-FLAIR & 140K & - & - & - & - & 60.5 & 58.2 & - \\
\midrule
\multicolumn{7}{l}{\textit{\textbf{32 frames, fixed 256$\times$28$\times$28 pixels per frame}}} \\
Qwen2.5-VL-7B-Instruct & - & 55.4 & 60.5	& 58.2 & 63.5 & 49.4 & 33.3 & 53.4 \\
Qwen3-VL-8B-Instruct & - & 62.6 & 65.3 & \underline{58.4} &  \underline{71.4} & \underline{63.8} & \underline{55.6} & \underline{62.9}\\
Video-R1 & 169K  & 59.3 & 63.9 & - & 63.8 & 52.3 & 35.8 & - \\
VideoRFT & 110K  & 59.8 & 62.1 & - & 68.5 & 51.1 & 36.8 & - \\
VideoAuto-R1-Qwen2.5 & 83K & 62.1 & \textbf{70.0} & 52.1 & 66.4 & 55.3 & 36.6 & 57.1 \\
VideoAuto-R1-Qwen3 & 83K & \textbf{63.4} & \underline{68.5} & 53.0 & 69.9 & 63.6 & 52.1 & 61.8 \\
\rowcolor{navyblue!10} EVAThink  & 6.7K & \underline{62.8} & 65.6 & \textbf{58.5} & \textbf{71.7} & \textbf{64.9}  & \textbf{57.2} & \textbf{63.5} \\
\rowcolor{navyblue!10}(think ratio)  & -  & (44.8\%) & (7.2\%) & (56.8\%) & (76.6\%) & (91.9\%) & (0.0\%) &   \\
\midrule
\multicolumn{7}{l}{\textit{\textbf{64 frames, fixed 256$\times$28$\times$28 pixels per frame}}} \\
Qwen2.5-VL-7B-Instruct & - & 62.6 & 67.5 & 60.1 & 64.5 & 55.1 & 38.6 & 58.1 \\
Qwen3-VL-8B-Instruct & -  & 64.9 & 65.8 & \underline{60.4} & \underline{70.4} & \underline{65.6} & \underline{58.3} & \underline{64.2} \\
VideoAuto-R1-Qwen2.5  & 83K & 64.1 & \textbf{69.8} & 53.3 & 67.0 & 55.1 & 39.2 & 58.1 \\
VideoAuto-R1-Qwen3 & 83K & \underline{66.0} & \underline{69.0} & 54.6 & 69.6 & 64.3 & 54.6 & 63.0 \\
\rowcolor{navyblue!10} EVAThink & 6.7K & \textbf{66.3} & 66.5 & \textbf{60.7} & \textbf{72.8} & \textbf{67.4} & \textbf{59.9} & \textbf{65.6} \\
\rowcolor{navyblue!10}(think ratio) &  & (34.4\%) & (5.1\%) & (46.1\%) & (59.4\%) & (85.7\%) & (0.0\%) &  \\
\bottomrule
\end{tabular}
}
\caption{Performance across video understanding benchmarks. \textbf{Bold} and \underline{underlined} values indicate the best and second-best scores within each fixed-budget group, respectively. Avg. is the unweighted mean over all six benchmarks and is reported only for complete results.
$^{\dagger}$AdaThinkV allows up to 256 input frames.}
\label{tab:exp_main}
\end{table*}

\section{Experiments}
\subsection{Experiment Setup}
\textbf{Training Setup.}
After the Recoverable Evidence Gap Mining, we obtain 1.76K high-confidence contrastive pairs (i.e., 3.5K samples) and 4.17K remaining contrastive pairs (i.e., 8.3K samples).
EVAThink is initialized from Qwen3-VL-8B-Instruct and trained on 32 Ascend 910B NPUs with a batch size of 32.
Stage 1 uses the high-confidence pairs for 110 steps, corresponding to approximately one epoch. Stage 2 continues training on the remaining pairs for 100 steps, using 3.2K samples. In total, the two stages use 6.7K training samples (See Appendix \ref{app:impl_detail_1} for more details).

\textbf{Baselines.}
We compare EVAThink with the Qwen2.5-VL-7B-Instruct \cite{bai2025qwen25vltechnicalreport} and Qwen3-VL-8B-Instruct \cite{bai2025qwen3}, as well as representative video reasoning methods: Temporal-RLT \cite{li2025reinforcement}, Video-RTS \cite{wang2025video}, VITAL \cite{zhang2026thinking}, LongVILA-R1 \cite{chen2026scaling}, LOVR-R1 \cite{fu2025love}, VideoChat-R1.5 \cite{yan2026videochat}, Video-R1 \cite{feng2026video}, VideoRFT \cite{wang2026videorft}, VideoAuto-R1 \cite{liu2026videoauto}, AdaThinkV \cite{tian2026adathinkv}, and Video-FLAIR \cite{kulkarni2026video}.
We distinguish results reported under the respective original evaluation protocols from those obtained under a controlled visual budget.
For the latter, we evaluate available baseline checkpoints using uniformly sampled 32 or 64 frames, with a fixed per-frame pixel budget of $256 \times 28 \times 28$. VideoAuto-R1 supports adaptive thinking and serves as a strong baseline evaluated under these same visual budgets.
For adaptive-thinking methods AdaThinkV and Video-FLAIR, we report published results under their original evaluation protocols because their checkpoints are unavailable.
Notably, AdaThinkV allows up to 256 frames, compared with our fixed budgets of 32 and 64.
In addition, we compare different response modes and ablate individual components to evaluate adaptive thinking and component contributions.
During evaluation, we remove EVAThink’s training-specific prompt.

\textbf{Benchmarks.}
We evaluate video understanding on six benchmarks: Video-MME \cite{fu2025video}, MVBench \cite{li2024mvbench}, LongVideoBench \cite{wu2024longvideobench}, MMVU (multiple-choice split) \cite{zhao2025mmvu}, Video-MMMU \cite{hu2026video}, and VSI-Bench \cite{yang2025thinking}. These benchmarks cover general and long-video understanding, as well as video perception and reasoning. We report accuracy on each benchmark and the unweighted average across all six, computing the average only when all results are available.
To assess zero-shot transfer to image understanding, we further evaluate on MMBench (MMBench-EN and MMBench-CN) \cite{liu2024mmbench}, MME (MME-P and MME-C) \cite{fu2026mme}, AI2D \cite{kembhavi2016diagram}, and POPE (accuracy) \cite{li2023evaluating} without additional image-task training.

\begin{table}[!t]
\centering
\renewcommand{\arraystretch}{1.1}
\setlength\tabcolsep{2.5em} 
\resizebox{\linewidth}{!}
{
\begin{tabular}{l|c|cc}
\toprule
\textbf{Model} & \textbf{Think Mode} &  \textbf{Avg.} & \textbf{Tok.} \\
\midrule
EVAThink & Direct & 63.3 & 8 \\
EVAThink & Think Only & 62.0 & 560 \\
\rowcolor{navyblue!10} EVAThink & Auto Think & 65.6 & 213 \\
\bottomrule
\end{tabular}
}
\caption{Effect of adaptive thinking under the 64-frame setting. Tok. denotes the average number of generated tokens per response.}
\label{tab:exp_thinkmode}
\end{table}

\begin{table}[!t]
\centering
\renewcommand{\arraystretch}{1}
\setlength\tabcolsep{1.5em}
\resizebox{\linewidth}{!}{%
\begin{tabular}{l|cc|c}
\toprule
\multirow{2}{*}{\textbf{Model}}
& \multicolumn{2}{c|}{\textbf{Think Ratio}}
& \multirow{2}{*}{\textbf{Overall Acc. (\%)}} \\
& Dense & Sparse & \\
\midrule
EVAThink & 22\% & 83\% & 65.3 \\
\bottomrule
\end{tabular}%
}
\caption{Evaluation on held-out dense/sparse pairs sharing the same question and answer, evaluated under the 64-frame setting.}
\label{tab:paired_evidence}
\end{table}

\begin{table*}[!thp]
\centering
\renewcommand{\arraystretch}{1.1}
\setlength\tabcolsep{2em} 
\resizebox{\linewidth}{!}
{
\begin{tabular}{l|cccccc|cc}
\toprule
\textbf{Model} & \textbf{MMBench-En} & \textbf{MMBench-Cn} &	\textbf{MME-P} & \textbf{MME-C} & \textbf{AI2D} & \textbf{POPE} & \multicolumn{2}{c}{\textbf{Rel.}}  \\
\midrule
Qwen3-VL-8B-Instruct & 83.7 & 82.2 & 1671.9 & 546.4 &  81.9 & 88.2 & \multicolumn{2}{c}{100\%}  \\
\rowcolor{navyblue!10} EVAThink & 85.1 & 82.9 & 1723.1 & 562.5 & 82.5 & 88.8 & \multicolumn{2}{c}{101.7\%} \\
\bottomrule
\end{tabular}
}
\caption{More results across image understanding tasks. “Rel.” denotes the average relative performance across benchmarks, computed as the mean percentage normalized by the base model (Qwen3-VL-8B-Instruct) set to 100\%.
}
\label{tab:exp_cross}
\end{table*}

\subsection{Main Results}
Table~\ref{tab:exp_main} shows performance on six video understanding benchmarks, including results under the original evaluation protocols and controlled comparisons with fixed budgets of 32 and 64 frames.
Under both fixed-budget settings, EVAThink achieves the highest average accuracy among the evaluated baselines with results on all six benchmarks. With 32 frames, EVAThink reaches 63.5\%, outperforming Qwen3-VL-8B-Instruct and VideoAuto-R1-Qwen3 by 0.6 and 1.7 percentage points, respectively. With 64 frames, its average accuracy increases to 65.6\%, extending these margins to 1.4 and 2.6 percentage points. These improvements are achieved using only 6.7K training samples, compared with 83K for VideoAuto-R1, demonstrating effective training with a small trainset.
Beyond the average gains, EVAThink consistently improves upon base model Qwen3-VL-8B-Instruct on every benchmark under both frame budgets, demonstrating benefits across video understanding tasks. 

Increasing the frame budget from 32 to 64 also reveals a consistent shift in response-mode selection.
Increasing the frame budget from 32 to 64 improves EVAThink's accuracy on all six benchmarks while reducing its thinking ratio on all five benchmarks where explicit reasoning is used.
This pattern is consistent with the motivation for evidence sufficiency-guided training: a larger frame budget may provide more question-relevant evidence and make direct answering more effective.
Meanwhile, the thinking ratio remains highest on Video-MMMU, consistent with its demands for concept understanding and knowledge application.
Conversely, the lowest thinking ratio on VSI-Bench suggests a preference for direct spatial judgments, aligning with prior findings \cite{yang2025thinking,kancheti2026chain,kulkarni2026video} that explicit verbal reasoning can hinder visual-spatial performance.

\subsection{Extended Results}
\textbf{Effect of adaptive thinking.}
We compare direct-answer, think-only, and default adaptive inference using the same EVAThink checkpoint under the 64-frame setting.
As shown in Table~\ref{tab:exp_thinkmode}, adaptive thinking achieves 65.6\% average accuracy, exceeding the two fixed modes by 2.3 and 3.6 percentage points, respectively, while reducing output length by 62\% relative to think-only inference.

We further evaluate response-mode selection on held-out dense/sparse pairs with matched questions and answers under the same 64-frame setting.
We randomly sample 200 crop-constructed pairs from the
evidence-contrastive pairs, excluding any pair whose source
video was used in training.
Each pair shares the same question and answer, with the
original video as the sparse view and an evidence-centered
crop as the dense view, evaluated under the same frame
budget and resolution.
EVAThink's thinking ratio increases from 22\% on dense views to 83\% on sparse views, while its overall accuracy reaches 65.3\%. These paired results suggest that EVAThink's mode selection is sensitive to evidence conditions, consistent with its training motivation.


\textbf{More results across image understanding tasks.}
To assess generalization beyond video understanding, we evaluate EVAThink on four image benchmarks without additional image-task training.
As shown in Table~\ref{tab:exp_cross}, beyond its consistent gains on video benchmarks, EVAThink also outperforms Qwen3-VL-8B-Instruct on every reported image-understanding metric, achieving an average relative performance of 101.7\%.
These results suggest that the benefits of EVAThink also transfer to image understanding.


\begin{table*}[!thp]
\centering
\renewcommand{\arraystretch}{1.1}
\setlength\tabcolsep{1em} 
\resizebox{\linewidth}{!}
{
\begin{tabular}{l|cccccc|cc}
\toprule
\textbf{Method} & \textbf{Video-MME} & \textbf{MVBench} &	\textbf{LongVideoBench} & \textbf{MMVU(mc)} & \textbf{Video-MMMU} & \textbf{VSI-Bench} & \textbf{Avg.} & \textbf{Tok.} \\
\midrule
w/o Recoverability filtering & 65.9 & 65.6 & 60.2 & 72.3 & 67.2 & 58.5 & 65.0 & 247 \\
w/o Evidence-gap \& Recoverability filtering & 64.8 & 65.1 & 59.5 & 69.8 & 63.3 & 58.2 & 63.5 & 198 \\
\midrule
w/o Stage 1 mode preference reward  & 63.2 & 62.7 & 56.4 & 68.2 & 59.7 & 56.7 & 61.2 & 392 \\
w/o Stage 2 direct-answer bonus reward & 66.0 & 65.9 & 60.5 & 72.6 & 68.1 & 58.6 & 65.3 & 465 \\
w/o First-token advantage regularization& 66.1 & 65.2 & 60.2 & 72.2 & 63.8 & 60.1 & 64.6 & 52 \\
\midrule
w/o Stage1 training & 66.2 & 66.8 & 60.0 & 71.5 & 63.2 & 58.9 & 64.4 & 80 \\
w/o Stage2 training & 65.4 & 65.2 & 61.3 & 73.4 & 66.7 & 60.3 & 65.4  & 331 \\
\midrule
\rowcolor{navyblue!10} EVAThink & 66.3 & 66.5 & 60.7 & 72.8 & 67.4 & 59.9 & 65.6 & 213 \\
\bottomrule
\end{tabular}
}
\caption{Ablation study under the 64-frame setting. RL optimization settings are held fixed; data-curation changes are described in the text.}
\label{tab:exp_ablation}
\end{table*}

\begin{figure}[!hbtp]
  \centering
  \includegraphics[width=0.495\linewidth]{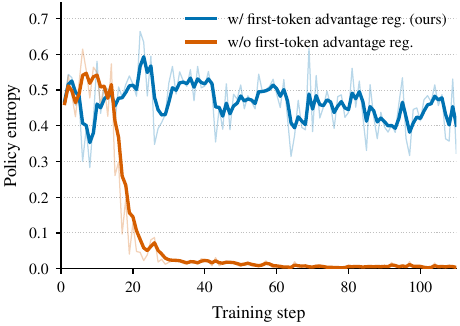}
  \hfill
  \includegraphics[width=0.495\linewidth]{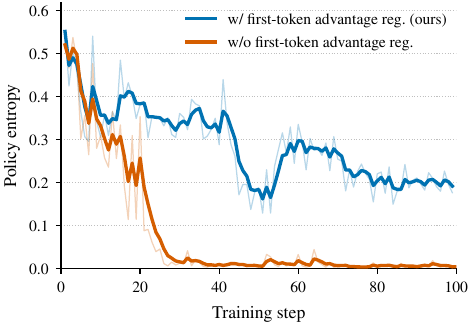}
  \caption{Effect of first-token advantage regularization on policy entropy during Stage~1 (left) and Stage~2 (right). Without regularization, entropy rapidly collapses in both stages, accompanied by a shift to direct answering for all rollouts (response-mode collapse).}
   \label{fig:entro}
\end{figure}

\begin{figure*}[!th]
  \centering
  \includegraphics[width=\linewidth]{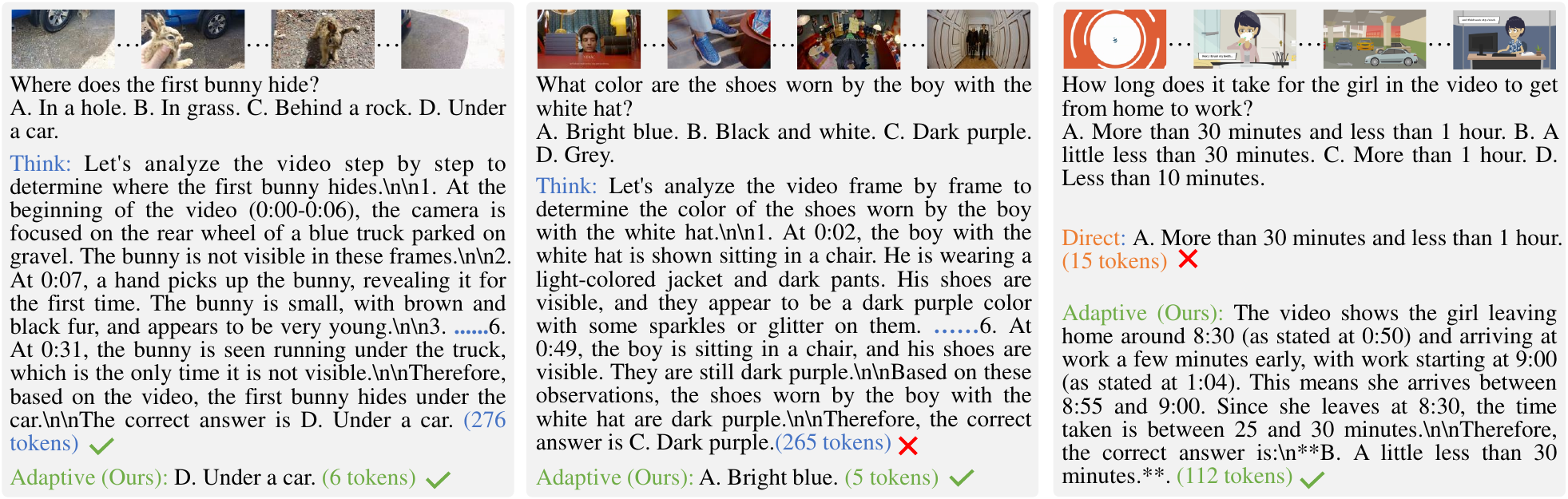}
  \caption{Case studies of adaptive thinking: EVAThink avoids unnecessary reasoning (left), prevents reasoning-induced errors through direct answering (middle), and resolves errors through evidence-grounded reasoning (right).}
   \label{fig:case_study}
\end{figure*}

\subsection{Ablation Study}
We ablate individual components of EVAThink under the 64-frame setting. Table~\ref{tab:exp_ablation} reports the resulting performance.

\noindent\textbf{Effect of data curation.}
We evaluate the effect of data curation by removing recoverability filtering alone and then removing both evidence-gap and recoverability filtering.
To accommodate the changes in the available training pairs, we adjust stage-wise sample selection accordingly: removing the $\Delta_{\mathrm{think}} \geq 0.5$ requirement for Stage~1 in the first variant, and randomly selecting samples for both stages from the candidate pairs in the second.
As shown in Table~\ref{tab:exp_ablation}, removing recoverability filtering reduces average accuracy from 65.6\% to 65.0\% while increasing average output length from 213 to 247 tokens, suggesting that identifying reasoning gains helps provide more effective supervision for adaptive thinking.
Removing both filters further reduces average accuracy to 63.5\%.
Although this variant produces slightly shorter responses (198 tokens), its lower accuracy indicates a less favorable accuracy-efficiency trade-off.
Together, these results support selecting evidence-contrastive pairs that exhibit both a direct-answer accuracy gap and a reasoning gain on the sparse view.

\textbf{Effect of reward design.}
We separately remove the Stage 1 mode-preference reward and the Stage 2 direct-answer bonus, using correctness-only answer rewards of +2/0 in Stage 1 and +1/0 in Stage 2 for correct/incorrect answers, respectively, while keeping all other objective components and training settings unchanged.
As shown in Table~\ref{tab:exp_ablation}, removing the Stage 1 mode-preference reward reduces average accuracy from 65.6\% to 61.2\% and increases average response length from 213 to 392 tokens, supporting its contribution to accurate and efficient adaptive answering.
Removing the Stage 2 direct-answer bonus yields comparable accuracy (65.3\% vs.\ 65.6\%) but substantially longer responses (465 vs.\ 213 tokens), indicating that the bonus improves response efficiency with little change in accuracy.

\textbf{Effect of first-token advantage regularization.}
Figure~\ref{fig:entro} illustrates the effect of first-token advantage regularization on training dynamics.
Without first-token advantage regularization, average accuracy decreases to 64.6\%, with the largest decline on Video-MMMU.
Its effect is also evident in the training dynamics: without regularization, policy entropy rapidly approaches zero in both stages, accompanied by mode collapse in which all rollouts shift to direct answering. In contrast, the regularized policy maintains nonzero entropy throughout training.
Together, these observations suggest that the regularizer helps prevent premature policy concentration and preserves exploration, contributing to better final performance.

\textbf{Effect of Stage~1 training.}
Omitting Stage~1 reduces average accuracy to 64.4\%, with particularly notable drops on Video-MMMU and MMVU of 4.2 and 1.3 points, respectively. The average output length also decreases to 80 tokens.
The combination of shorter responses and lower accuracy suggests that the model may invoke reasoning less effectively without Stage~1.
These results support the role of Stage~1 in establishing an adaptive response-mode policy before Stage~2 refines response efficiency.

\textbf{Effect of Stage~2 training.}
Omitting Stage~2 yields an average accuracy of 65.4\%, slightly below the full model, while increasing the average output length from 213 to 331 tokens. Although Stage~2 does not improve every benchmark, it raises overall accuracy and reduces output length by approximately 36\% relative to the Stage~1-only model. These results suggest that Stage~2 refines the learned policy, improving the overall balance between answer accuracy and reasoning length.

\subsection{Case Study}
Figure~\ref{fig:case_study} presents examples of EVAThink's adaptive thinking. On the left, EVAThink answers correctly without explicit reasoning, using 6 tokens compared with 276 in think-only mode. In the middle, it correctly identifies the shoe color through direct answering, whereas think-only mode answers incorrectly. On the right, it uses visual time cues to correctly infer a duration that direct-answer mode gets wrong. These cases illustrate how adaptive thinking can reduce unnecessary computation, avoid reasoning-induced errors, and improve answers through evidence integration.

\section{Conclusion}
We presented EVAThink, which uses evidence sufficiency to guide adaptive-thinking training for video understanding.
Recoverable Evidence Gap Mining constructs evidence-contrastive pairs to supervise mode selection, while two-stage reinforcement learning and first-token advantage regularization support effective and efficient adaptive thinking. Using only 6.7K training samples, EVAThink improves video understanding under fixed visual budgets. Compared with always reasoning, it achieves higher average accuracy with substantially shorter responses, supporting evidence sufficiency-guided training for adaptive thinking.
\clearpage
{
    \small
    \bibliographystyle{ieeenat_fullname}
    \bibliography{main}
}

\clearpage
\appendix
\setcounter{page}{1}

\twocolumn[
  \begin{center}
    \Large \textbf{Appendix}
    \vspace{0.5cm} 
  \end{center}
]

\section{More Details of Data Pool}
\label{app:data_pool}
We construct the data pool of approximately 118.2K samples from eight video understanding datasets. Table~\ref{tab:app_compo} summarizes the data sources, selection criteria, and corresponding sample counts. Specifically, we use the academic subset of LLaVA-Video-178K and randomly sample subsets from Molmo2-VideoCapQA and VideoMarathon. For SR-91K, we retain the multiple-choice subset. We include all samples from LongVILA, Holmes-train, NeXT-QA, and PerceptionTest. These sources form the candidate pool for subsequent contrastive-pair construction and filtering.

\begin{table}[htp]
\centering
\renewcommand{\arraystretch}{1.1}
\setlength\tabcolsep{1em} 
\resizebox{\linewidth}{!}
{
\begin{tabular}{l|cc}
\toprule
\textbf{Data source} & \textbf{Selection criterion} & \textbf{\# Selected samples}  \\
\midrule
LLaVA-Video-178K & Academic subset & 28.9K   \\
Molmo2-VideoCapQA & Random subset & 21.9K \\
VideoMarathon & Random subset & 4.5K \\
LongVILA & All & 29.1K \\
SR-91K & Multiple-choice subset & 18.4K \\
Holmes-train & All & 1.6K \\
NeXT-QA & All & 7.5K \\
PerceptionTest & All & 6.3K \\
\bottomrule
\end{tabular}
}
\caption{Composition of the data pool.
}
\label{tab:app_compo}
\end{table}

\section{Implementation Details of EVAThink}
\label{app:impl_detail_0}
\subsection{Prompts for Rollouts}
Tables~\ref{tab:app_prompt_direct} and~\ref{tab:app_prompt_think} present the prompts used for direct-answer and thinking rollouts during data curation. The former requests a direct answer, while the latter instructs the model to reason over cross-frame clues before answering, without introducing unsupported details. Both prompts require the final answer to be enclosed in \texttt{<answer></answer>} tags, containing only the option letter for multiple-choice questions.

\begin{table}[th]
    \centering
    \footnotesize
    \resizebox{0.99\linewidth}{!}{
    \begin{tabular}{p{\linewidth}}
        \toprule
        {\scriptsize\textbf{\textit{USER PROMPT FOR DIRECT ANSWER}}} \\
        \midrule

You are given a video. Please answer the question directly.

Question:
\texttt{[Question]}

\texttt{[Options]}

Your final answer must be enclosed in \texttt{<answer></answer>}.

- For a multiple-choice question, put only the option letter inside \texttt{<answer></answer>}.

Examples:
\texttt{<answer>}A\texttt{</answer>} \\
    \bottomrule
\end{tabular}}
\caption{User prompt for direct answer in the data curation.}
\label{tab:app_prompt_direct}
\end{table}

\begin{table}[!ht]
    \centering
    \footnotesize
    \resizebox{0.99\linewidth}{!}{%
    \begin{tabular}{p{\linewidth}}
        \toprule
        {\scriptsize\textbf{\textit{USER PROMPT FOR THINKING}}} \\
        \midrule
        You are given a video. \\
        Do not answer immediately. Think step by step before answering. \\
        If the frames are temporally jumpy, the evidence is scattered, or the visible content is not sufficient for an immediate answer, carefully reason over the observed cross-frame clues and integrate the evidence before giving your answer. \\
        Do not invent details that are not supported by the visible frames. \\[4pt]
        \textbf{Note:} \\
        Work through your reasoning first, then give your final answer. \\
        \textbf{Your final answer MUST be enclosed in \texttt{<answer></answer>}.} \\
        \textbf{For multiple-choice questions, put only the option letter inside the tags, e.g., \texttt{<answer>A</answer>}.} \\[4pt]
        \textbf{An Example:} \\
        Question: What does the person add to the bowl? \\
        Options: \\
        A. Water \\
        B. Milk \\
        C. Flour \\
        D. Oil \\[4pt]
        The sampled frames are not fully continuous, so I need to connect the visible clues across time. In an earlier frame, the person is holding a carton near the bowl. In a later frame, a white liquid is being poured into the bowl. The final frames show the bowl containing a white liquid. I do not see evidence of flour, oil, or clear water being added. The scattered visual clues together support that the person adds milk. \\[4pt]
        \texttt{<answer>B</answer>} \\[4pt]
        \texttt{===} \\[4pt]
        Now answer the actual question: \\[4pt]
        Question: \\
        \texttt{[Question]} \\[4pt]
        Options: \\
        \texttt{[Options]} \\
        \bottomrule
    \end{tabular}}
    \caption{User prompt for thinking in the data curation.}
    \label{tab:app_prompt_think}
\end{table}

\subsection{Details of Evidence Segment Annotation}

\begin{table*}[!t]
    \centering
    \footnotesize
    \resizebox{0.99\linewidth}{!}{%
    \begin{tabular}{p{\linewidth}}
        \toprule
        {\scriptsize\textbf{\textit{USER PROMPT}}} \\
        \midrule
        You are a video evidence localizer. \\
        You will be given a video constructed from exactly 32 frames uniformly sampled from the original video and then arranged in chronological order. \\
        The frame ids are defined by their order in this 32-frame video: \\[4pt]
        - The first sampled frame is Frame 0. \\
        - The last sampled frame is Frame 31. \\[4pt]
        Your task is not to answer the question. \\
        Your task is to locate the sampled frame range that contains the visual evidence needed to support the given ground-truth answer. \\[4pt]
        Question: \\
        \texttt{[Question]} \\[4pt]
        Options: \\
        \texttt{[Options]} \\[4pt]
        Ground-truth answer: \\
        \texttt{[Answer]} \\[4pt]
        Rules: \\
        1. Output the minimal sampled frame range that visually supports the ground-truth answer. \\
        2. If the required evidence appears in multiple separated moments, output the smallest continuous range covering all required evidence. \\
        3. Use frame ids (starting from 0), not estimated timestamps. \\
        4. Do not include irrelevant context unless it is needed to answer the question. \\
        5. If the evidence is not visible in the provided frames, output \texttt{status = "not\_localized"}; if the evidence is ambiguous or only partially visible, output \texttt{status = "partial"}. \\
        6. Both \texttt{start\_frame} and \texttt{end\_frame} are included in the selected range. \\[4pt]
        Return only valid JSON in the following format: \\[4pt]
        \texttt{\{} \\
        \texttt{\ \ "status": "localized | partial | not\_localized",} \\
        \texttt{\ \ "start\_frame": int,} \\
        \texttt{\ \ "end\_frame": int,} \\
        \texttt{\ \ "confidence": "high | medium | low"} \\
        \texttt{\}} \\
        \bottomrule
    \end{tabular}}
    \caption{Prompt for video evidence segment annotation, which instructs the model to identify the minimal sampled frame range supporting the ground-truth answer.}
    \label{tab:app_prompt_anno}
\end{table*}

We use a stronger model to annotate evidence segments (the prompt is shown in the Table \ref{tab:app_prompt_anno}). Given 32 uniformly sampled frames in chronological order, together with the question, candidate options, and ground-truth answer, the annotator identifies the minimal continuous frame interval containing the supporting evidence. When the required clues occur at separate moments, the interval covers all of them. Annotations use inclusive frame indices and include a localization status and confidence level.
Ambiguous or absent evidence is explicitly marked as \texttt{partial} or \texttt{not\_localized}, respectively, and the corresponding samples are discarded.

\noindent\textbf{Annotation accuracy.}
The annotations serve as candidate evidence regions for mining, so precise temporal boundaries are not required. Sample acceptance is subsequently determined by two filtering stages based on rollout accuracy, which verify whether each candidate exhibits the behavior required by our mining criteria. These checks reduce sensitivity to localization errors: an annotation alone is insufficient for a sample to be retained. Thus, the retained samples satisfy our operational criteria for recoverable evidence gaps, without requiring frame-exact annotations.

\noindent\textbf{Annotation cost.}
Although annotation uses a stronger model, it is applied only to a small subset of the original data pool. After removing samples for which all preceding rollouts are correct, only the remaining samples identified as sparse-view cases require evidence annotation. Empirically, these account for less than 22.7\% of the original pool. Each annotation call receives only 32 sampled frames and requests a compact localization output, keeping the overall annotation budget controlled.

\subsection{Details of Resulting Evidence-Contrastive Pairs}

Table~\ref{tab:app_trainset} summarizes the evidence-contrastive pairs retained from each data source for the two training stages. The resulting training sets contain 1,766 pairs for Stage 1 and 4,179 pairs for Stage 2, spanning eight video datasets. Both stages include pairs from all eight sources, with SR-91K contributing the largest share in each stage.

\begin{table}[htp]
\centering
\renewcommand{\arraystretch}{1.1}
\setlength\tabcolsep{1em} 
\resizebox{\linewidth}{!}
{
\begin{tabular}{l|cc}
\toprule
\textbf{Data source} & \textbf{Stage 1 (pairs)} & \textbf{Stage 2 (pairs)}  \\
\midrule
LLaVA-Video-178K & 317 & 628   \\
Molmo2-VideoCapQA & 325 & 529 \\
VideoMarathon & 51 & 169 \\
LongVILA & 329 & 720 \\
SR-91K & 514 & 1682 \\
Holmes-train & 28 & 67 \\
NeXT-QA & 75 & 136 \\
PerceptionTest & 127 & 248 \\
\midrule
Total & 1766 & 4179 \\
\bottomrule
\end{tabular}
}
\caption{Composition of the evidence-contrastive training pairs for the two stages.
}
\label{tab:app_trainset}
\end{table}

\begin{table*}[th]
    \centering
    \footnotesize
    \resizebox{0.99\linewidth}{!}{
    \begin{tabular}{p{\linewidth}}
        \toprule
        {\scriptsize\textbf{\textit{USER PROMPT}}} \\
        \midrule

        Answer the question based on the video. First decide whether explicit written reasoning is necessary. Choose exactly one of the following two response modes. Your response must begin with either \texttt{<answer>} or \texttt{Alright} with no other opening. \\

        \textbf{1. Direct-answer mode} \\
        Use this mode when the visual evidence clearly supports the final answer and no written reasoning is needed. Begin directly with \texttt{<answer>} and output only the final answer block. \\
        Example: \texttt{<answer>}A\texttt{</answer>} \\

        \textbf{2. Evidence-grounded reasoning mode} \\
        Use this mode only when the answer is not immediately clear, or when the relevant evidence is temporally sparse, discontinuous, or scattered across multiple frames. Begin with \texttt{Alright}, briefly connect the relevant observed clues, and then provide the final answer block. Include only as much reasoning as necessary to determine the answer. \\
        Example: \texttt{Alright}, the earlier frames show the person picking up the cup, and the later frames show them drinking from it... \texttt{<answer>}A\texttt{</answer>} \\

        \textbf{For every response:} \\
        - The response must begin with either \texttt{<answer>} for direct answering or \texttt{Alright} for evidence-grounded reasoning. Do not use any other opening. \\
        - Put the final answer only inside \texttt{<answer>}A\texttt{</answer>} at the very end of the response. \\
        \bottomrule
    \end{tabular}}
    \caption{Training prompt for response mode selection, which instructs the model to adaptively select between direct and thinking modes.}
    \label{tab:prompt_training}
\end{table*}

\begin{table*}[!th]
\centering
\renewcommand{\arraystretch}{1.1}
\setlength\tabcolsep{1em} 
\resizebox{\linewidth}{!}
{
\begin{tabular}{l|c|cccccc|cc}
\toprule
\textbf{Model} & \textbf{Think Mode} & \textbf{Video-MME} & \textbf{MVBench} &	\textbf{LongVideoBench} & \textbf{MMVU(mc)} & \textbf{Video-MMMU} & \textbf{VSI-Bench} & \textbf{Avg.} & \textbf{Tok.} \\
\midrule
EVAThink & Direct & 66.2 & 66.4 & 59.8 & 70.6 & 56.8 & 59.9 & 63.3 & 8 \\
EVAThink & Think Only & 60.8 & 64.2 & 57.4 & 71.2 & 67.1 & 51.4 & 62.0 & 560 \\
\rowcolor{navyblue!10} EVAThink & Auto Think & 66.3 & 66.5 & 60.7 & 72.8 & 67.4 & 59.9 & 65.6 & 213 \\
\rowcolor{navyblue!10}(think ratio) &  & (34.4\%) & (5.1\%) & (46.1\%) & (59.4\%) & (85.7\%) & (0.0\%) & & \\
\bottomrule
\end{tabular}
}
\caption{Effect of adaptive thinking under the 64-frame setting. Tok. denotes the average number of generated tokens per response.}
\label{tab:app_exp_thinkmode}
\end{table*}

\subsection{Training Prompt}
\label{app_training_prompt}
Table~\ref{tab:prompt_training} presents the training prompt, which instructs the model to select between direct answering and evidence-grounded reasoning based on the available visual evidence. Direct responses begin with \texttt{<answer>}, while reasoning responses begin with \texttt{Alright} and briefly connect observed clues before answering. Both modes place the final answer within \texttt{<answer></answer>} tags. The distinct openings allow the selected response mode to be identified directly from the output.

\section{Details of Experiments}
\label{app:impl_detail_1}
\subsection{Training Hyperparameters}
Table~\ref{tab:app_hyperparam} summarizes the training hyperparameters. We use AdamW with a constant learning rate of $10^{-6}$ and a batch size of 32. For each prompt, we sample eight rollouts at a temperature of 1.0 and perform one policy update per step. Gradient clipping is set to 1.0, and the maximum prompt and response lengths are 20,480 and 1,024 tokens, respectively.

\begin{table}[htbp]
\centering
\renewcommand{\arraystretch}{1.1}
\setlength\tabcolsep{1em} 
\resizebox{0.6\linewidth}{!}
{
\begin{tabular}{l|c}
\toprule
\textbf{Hyperparameter} & \textbf{Value} \\
\midrule
Batch size & 32   \\
Rollouts per prompt & 8 \\
Policy updates per step & 1 \\
Optimizer & AdamW\\
Learning rate & 1e-6 (constant) \\
Gradient clipping & 1.0 \\
Rollout temperature & 1.0 \\
Max prompt length & 20480 \\
Max response length & 1024 \\
\bottomrule
\end{tabular}
}
\caption{Training Hyperparameters.}
\label{tab:app_hyperparam}
\end{table}

\subsection{Evaluation Settings}
All evaluations use greedy decoding with the temperature set to 0.
The maximum response length is set to 2048 tokens.

\subsection{More Results of Effect of Adaptive Thinking}
Table~\ref{tab:app_exp_thinkmode} provides per-benchmark results for direct-answer, think-only, and adaptive inference using the same EVAThink checkpoint under the 64-frame setting.
Adaptive inference achieves 65.6\% average accuracy with 213 output tokens per response, compared with 63.3\% and 8 tokens for direct answering and 62.0\% and 560 tokens for think-only inference.
It matches or exceeds both fixed modes on every benchmark.





\section{Case Study}
\begin{figure*}[!th]
  \centering
  \includegraphics[width=\linewidth]{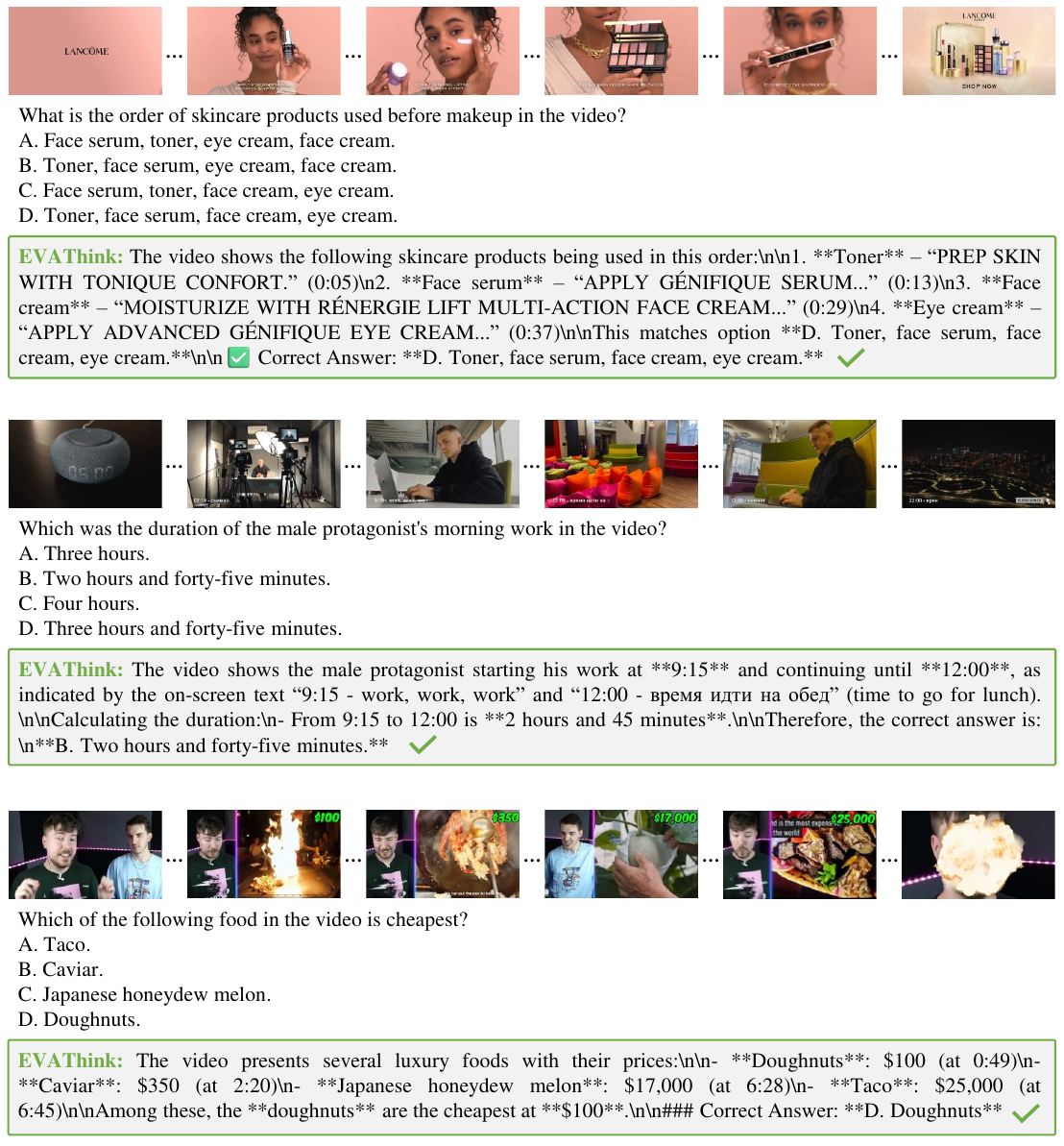}
  \caption{Case studies of EVAThink: think mode (1).}
   \label{fig:app_case_think_1}
\end{figure*}

\begin{figure*}[!th]
  \centering
  \includegraphics[width=\linewidth]{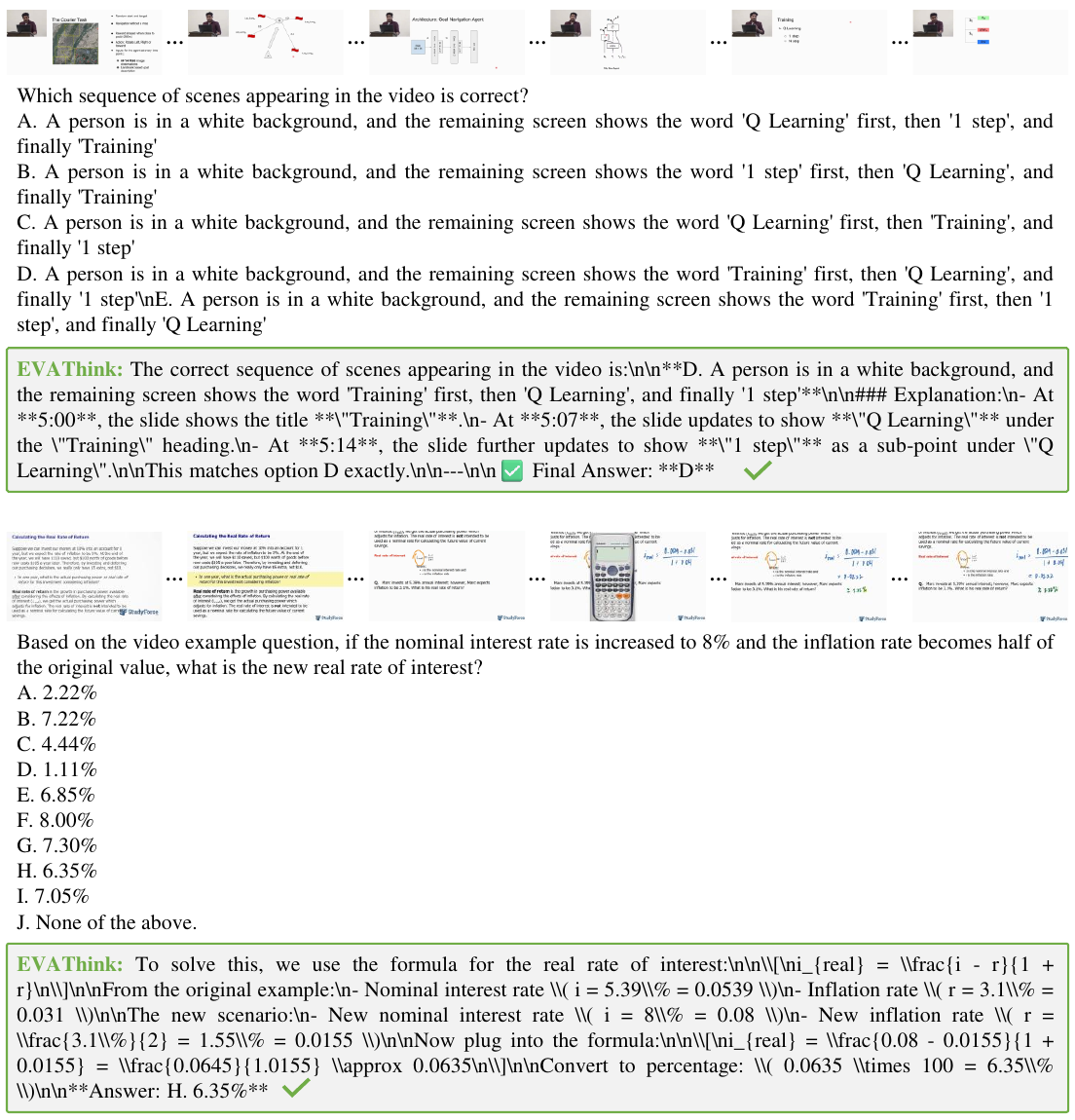}
  \caption{Case studies of EVAThink: think mode (2).}
   \label{fig:app_case_think_2}
\end{figure*}

\begin{figure*}[!th]
  \centering
  \includegraphics[width=\linewidth]{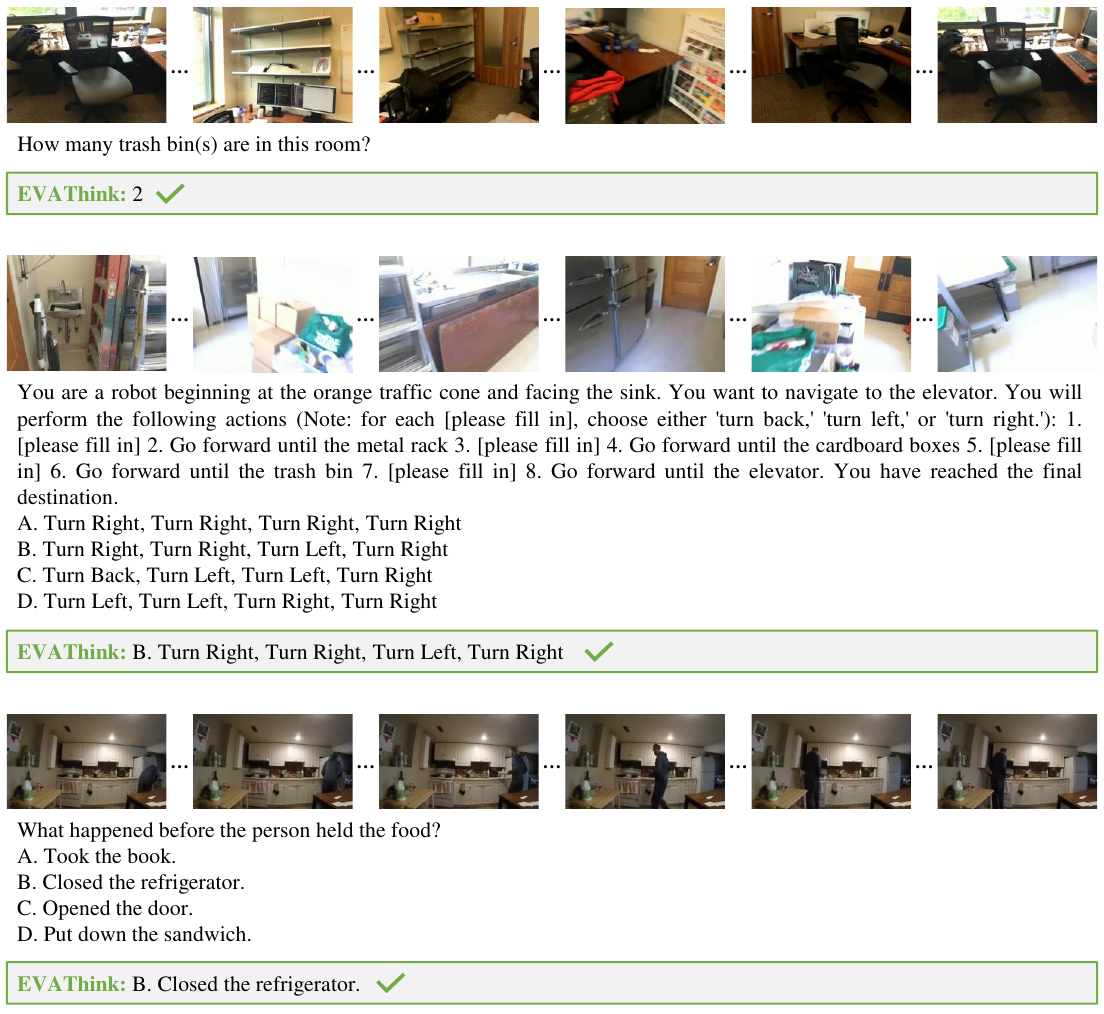}
  \caption{Case studies of EVAThink: direct mode (1).}
   \label{fig:app_case_direct_1}
\end{figure*}

\begin{figure*}[!th]
  \centering
  \includegraphics[width=\linewidth]{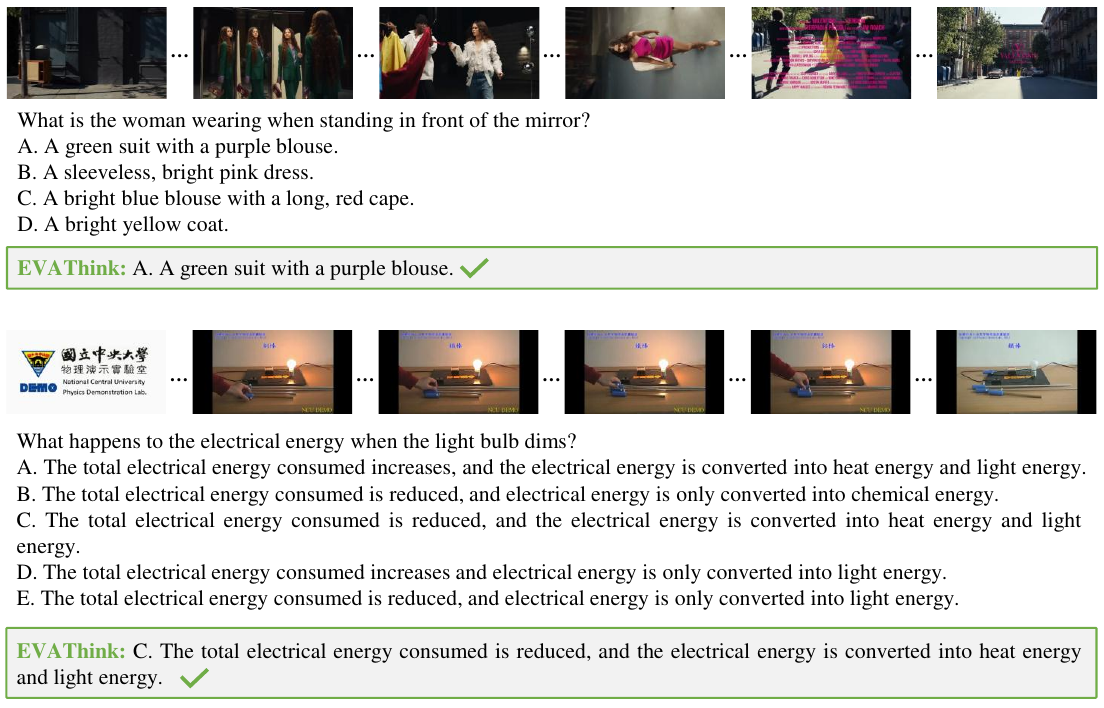}
  \caption{Case studies of EVAThink: direct mode (2).}
   \label{fig:app_case_direct_2}
\end{figure*}

Figures~\ref{fig:app_case_think_1} and~\ref{fig:app_case_think_2} present examples of EVAThink in thinking mode, while Figures~\ref{fig:app_case_direct_1} and~\ref{fig:app_case_direct_2} present examples in direct-answer mode. These cases illustrate the two response modes, showing how the model connects visual clues through explicit reasoning or provides an answer directly.

\clearpage
\clearpage
\clearpage
\clearpage

\end{document}